\relax
\documentclass[letterpaper]{article} % DO NOT CHANGE THIS
\usepackage{aaai20}  % DO NOT CHANGE THIS
\usepackage{times}  % DO NOT CHANGE THIS
\usepackage{helvet} % DO NOT CHANGE THIS
\usepackage{courier}  % DO NOT CHANGE THIS
\usepackage[hyphens]{url}  % DO NOT CHANGE THIS
\usepackage{graphicx} % DO NOT CHANGE THIS
\usepackage{graphicx}  % DO NOT CHANGE THIS
\usepackage{url}
\usepackage{booktabs}
\usepackage{amsmath}
\usepackage{lipsum}% http://ctan.org/pkg/lipsum
\usepackage{graphicx}% http://ctan.org/pkg/graphicx
\usepackage{subcaption}
\usepackage{caption}
\usepackage{xcolor}

\usepackage{amsmath}
\usepackage{amssymb}

\title{Improving Conditioning in Context-Aware Sequence to Sequence Models}
\author{Xinyi Wang,\\
{LTI, Carnegie Mellon University\thanks{Work done while interning at FAIR NY}}\\
xinyiw1@cs.cmu.edu
\And
Yacine Jernite, Jason Weston, Michael Auli\\
{Facebook AI Research}\\
\{yjernite, jase, michaelauli\}@fb.com}

\begin{document}

\maketitle

\begin{abstract}
Neural sequence to sequence models are well established for applications which can be cast as mapping a single input sequence into a single output sequence. In this work, we focus on cases where generation is conditioned on both a short query and a long context, such as abstractive question answering or document-level translation. We modify the standard sequence-to-sequence approach to make better use of both the query and the context by expanding the conditioning mechanism to intertwine query and context attention. We also introduce a simple and efficient data augmentation method for the proposed model. Experiments on three different tasks show that both changes lead to consistent improvements.
% Neural sequence to sequence models have recently lead to significant advances in a number of natural language processing tasks, such as machine translation or text summarization. However, while the paradigm is well established for applications that transform a single text into a single output sequence, the question of how best to adapt the method to composite inputs remains open. In this work, we focus on cases where generation is conditioned on both a shorter query sequence and a longer context, such as machine translation with document-level context or abstractive question answering conditioned on evidence documents.
% We modify the standard sequence to sequence approach to help the model make better use of both the query and context in two ways. First, we expanding the conditioning mechanism to intertwine query and context attention. Second, we introduce a simple and efficient data augmentation method for the proposed model. We validate our methods experimentally on three different tasks and show that both changes contribute to consistent improvements across all considered settings.
\end{abstract}
\section{\label{sec:intro}Introduction}
%\begin{figure*}
%    \centering
%    \includegraphics[width=\textwidth]{figs/example_eli5.pdf}
%    \caption{Caption}
%    \label{fig:my_label}
%\end{figure*}

\begin{table*}[!t]
    \centering
    \resizebox{!}{0.15\textwidth}{
    \begin{tabular}{p{3cm}|p{5cm}|p{10.1cm}|p{3.8cm}}
    \toprule
    Task & Source~($S$) & Context~($C$) & Target~($T$) \\
    \midrule
    Long Form QA & The difference between adverbs and adverbials. & Specifically, adverbs tell us about\ldots Some of what follows will cover some adverbs but in the context of different types of adverbial rather than as a distinct and discrete word class. The point to remember is: adverbs are all adverbials but adverbials are not all adverbs. Very briefly: He went yesterday contains an adverb, yesterday, acting as an adverbial telling us when he left\ldots The trick, as always, is to look at what the word is doing, not what it looks like. It is not only adverbs that do this\ldots & An adverb is a word, an adverbial is a group of words, phrase or clause which acts as an adverb. \\
    \midrule
    Knowledge-Grounded Dialogue & the eight members are brown university, columbia university, cornell university, dartmouth college, harvard university, the university of pennsylvania, princeton, and yale. & i grew up in virginia, also known as it's nickname "old dominion". really? i have never been, is it a interesting place? it is, you can check out old dominion university. its campus covers neiborhoods inculding highland park and downtown norfolk. ok is it considered an ivy league school? & i don't think so, i think there are only 8 including brown, columbia, cornell, dartmouth, harvard and a few others. \\
    \midrule
    Document-Level NMT & d\`es lors, choisiriez vous le m\^eme type de vacances? & je voudrais que vous pensiez à une expérience c\'er\'ebrale. imaginez qu'\`a vos prochaines vacances, qu' \`a la fin de ces vacances, toutes vos photos soient d\'etruites, et que vous avaliez un comprim\'e qui rend amn\'esique de tel sorte que vous ne vous souviendrez de rien. & now, would you choose the same vacation? \\
    \bottomrule
    \end{tabular}}
    \caption{Example of Source~(S), Context~(C), and Target~(T) for three context-aware sequence-to-sequence tasks.}
    \label{tab:examples}
\end{table*}

Many traditional natural language processing tasks can be formulated as a sequence transduction problem, where a system is given an input text sequence and asked to provide a corresponding natural language output.
In recent years, the development of neural sequence-to-sequence (seq2seq) models with attention mechanisms has lead to significant progress on such tasks.
These include for example machine translation when mapping text from one language to another, dialogue agents which read the last few utterances and produce a next chat, or text summarization systems.

The success of these models in context-free seq2seq applications has paved the way for systems which consider more complex inputs. For example, while sentence-level machine translation models have made great strides, they struggle at handling cases where the meaning of a word depends on the previous context in the document being translated. %~\cite{review_ctx_nmt}.
Similarly, while question answering can be formulated as a seq2seq problem, it also usually relies on context provided in the form of one or many supporting documents containing external knowledge.

In this work, we consider the problem of adapting the traditional seq2seq approach to the context-aware setting, where a model is given both a source and some context containing relevant information and required to produce the corresponding output. First, we explore different ways of combining the source and context when decoding the output, including alternating and interleaving attention layers. We then present a data augmentation technique which takes advantage of the proposed models' structure.

We apply our approach to three context-aware seq2seq tasks: neural machine translation with document-level context, long form question answering, where the system needs to provide a paragraph-length answer to a complex question given a support document, and knowledge-grounded dialogue, where the system is given a short fact and a dialogue history, and required to provide a next dialogue utterance given the provided fact. We show that our models consistently improve over standard seq2seq modeling as well as strong baselines for each of the tasks by helping the model make better use of both source and context.

% However, while the paradigm is straight-forward to apply to and has been demonstrated to work well in cases where the input consists of a single sequence, many language tasks consider more complex inputs. In particular, a number of applications rely on a longer context text sequence to answer a shorter query. In this work, we consider how to adapt the standard neural sequence-to-sequence model to such settings; namely, abstractive question answering with a supporting document, machine translation with document-level context, and knowledge-grounded dialogue, where the knowledge is represented as a text document.

\section{Related Work}

The development of neural seq2seq models with attention, which were first applied to the task of sentence-level machine translation~\cite{nmt_bahdanau,nmt_luong}, has lead to significant advances in several other natural language processing tasks which can be formulated as producing a sentence or paragraph output given a single text input. These include for example summarization~\cite{summ_rush,summ_see}, or  dialogue agents which take a chat history as input to produce a next utterance~\cite{dial_e2e}. Following the success of recurrent~\cite{nmt_bahdanau} and convolutional seq2seq~\cite{gehring2017convs2s} approaches, the Transformer architecture which generalizes the use of attention to the encoder and decoder has recently lead to further improvements on these applications~\cite{transformer}.

In this work, we consider tasks which, in addition to the main input sequence, rely on external information provided in the form of a context document. 
For example, abstractive question answering tasks make use of information found in web pages~\cite{msmarco,eli5}, or even full books~\cite{narrative_qa}, and the WikiSum dataset~\cite{wikisum} challenges a system to write a Wikipedia article given a title and a set of supporting web documents.
This family of problems also includes a versions of machine translation which looks at both the current sentence and past document history, as leveraging contextual information beyond the sentence being considered to improve translation is an outstanding challenge~\cite{review_ctx_nmt,doc_nmt}.
Finally, knowledge-grounded dialogue systems require an agent to produce a next dialogue utterance conditioned both on the chat history and on facts found e.g. in Wikipedia articles~\cite{wizard}.

The most straight-forward approach to applying encoder-decoder seq2seq models to these problems is to simply concatenate the source and context sequences and treating them as a single input. However, \cite{wikisum} find that this approach struggles with handling longer contexts. \cite{eli5} obtain some promising results using this architecture, but their approach requires a computation intensive multitask training strategy.

One of the main challenges of these tasks is the length of the context relative to the source sequence. Approaches to addressing this issue have included memory networks~\cite{e2e_memnet}, or combining recurrent and attention-based representations at different levels to incorporate longer context in language models~\cite{longlm,transformer_xl}. For contextual machine translation, recent works have also proposed using a memory cache~\cite{tu_cache} or hierarchical document representation to encode additional context information~\cite{hier_doc_nmt}.

\section{Context-Aware Sequence-to-Sequence Tasks}

In this Section, we start by formalizing the problem of context-aware sequence to sequence modeling and outlining some of its inherent challenges. We then review the Transformer encoder-decoder architecture~\cite{transformer}, propose structural modifications to adapt it to the context-aware setting, and introduce a data augmentation method to help train the proposed models more efficiently.

\subsection{\label{sec:problem}Problem Description}

In this work, we consider the family of context-aware sequence to sequence tasks. In the standard seq2seq setting, the goal is to generate an answer or target sequence ${T=\{t_1, t_2, \ldots, t_p\}}$ with $p$ tokens given a query or source sequence ${S=\{s_1, s_2, \ldots, s_m\}}$ with $m$ tokens. In the context-aware setting, the model is also required to make use of an additional sequence ${C=\{c_1, c_2, \ldots, c_n\}}$ which contains information relevant to the query (or source) $S$.

Proper utilization of both the query $S$ and context $C$ introduces several challenges. On the one hand, context sequences tend to be much longer and  in some cases noisier than the queries. For example, in the case of long form QA, the context sequences are made up of passages of Internet pages coming from all sorts of domains that could be relevant for answering the question~\cite{eli5}. Each context sequence is a concatenation of potentially (but not always) useful information segments and can be as long as 1000 words. Therefore, the encoding of the context $C$ should allow the model to focus on the most important information.

On the other hand, ensuring that the model produces outputs that are relevant to the query $S$, rather than generic answers has been shown to be challenging in conditional text generation tasks~\cite{ctx_sensitive,diverse}. Additionally, in the context-aware setting, we ask the model not only to produce an output that is germane to the input, but also to take into account the relative importance of the source and context when writing it (in most cases, relying chiefly on the source $S$ and using $C$ as secondary support).

Finally, collecting data for context-aware sequence to sequence tasks can be more difficult than for context-free applications. For question answering, gathering relevant supporting documents can represent a significant investment~\cite{msmarco}. Even for machine translation where context-aware data is relatively easy to collect, most of the standard datasets available now are still context-free~\cite{review_ctx_nmt,doc_nmt}. This makes model over-fitting even more of a concern in this setting. 

\subsection{Transformer Sequence-to-Sequence Model}

Based on the challenges outlined above and drawing inspirations from recent advances seq2seq modeling, we propose several models for context-aware seq2seq problems. Our architectures are based on the Transformer encoder-decoder model~\cite{transformer}; we review both their encoder and decoder architecture next.

\paragraph{Attention} The main component of the Transformer model is an attention module $\text{attn}(\cdot, \cdot, \cdot)$ which takes as input a set of $N$ query vectors of size $D$, as well as a set of $M$ key and value vectors: $Q \in \mathbb{R}^{N \times D}$, $K \in \mathbb{R}^{M \times D}$, and $V \in \mathbb{R}^{M \times D}$ respectively. For each query vector, the module computes an attention distribution $\alpha$ over the key vectors:
\begin{align}
    \label{eqn:attn_score}
    \alpha_{Q, K} = \text{softmax} \Big( (Q P_Q) (K P_K)^T \Big)
\end{align}
where $P_Q \in \mathbb{R}^{D \times D}$ and $P_K \in \mathbb{R}^{D \times D}$ are model parameters. The module then returns a corresponding convex combination of the value vectors:
\begin{align}
    \label{eqn:attn_return}
    \text{attn}(Q, K, V) = \alpha_{Q, K} V
\end{align}

\paragraph{Encoder} The Transformer encoder consists of $L_E$ identical layers, where each layer consists of a self-attention module ($\text{attn}(\cdot, \cdot, \cdot)$) which uses the output of the previous layer as queries, keys, and values, followed by a feed forward module ($FF(\cdot)$). That is:
    \begin{align}
        \label{eqn:enc_layers}
        \forall l \in \{1\ldots L_E\}, \quad A_l^E = FF\Big(\text{attn}(A_{l-1}^E, A_{l-1}^E, A_{l-1}^E)\Big)
    \end{align}
Where the first layer attends over the initial encoding of the source sequence: $A_0^E=S$.

\paragraph{Decoder} The decoder architecture is similar to the encoder, with two major differences. First, since the model needs to be used to decode the output sequence one word at a time, the self-attention for each time step can only be computed over previous time steps. We denote this version of the attention as $\overline{\text{attn}}(\cdot, \cdot, \cdot)$. Secondly, the decoder layers add an attention module over the encoder output between the self-attention and feed-forward modules. That is, for a decoder with $L_D$ layers, $\forall l \in \{1 \ldots L_D\}$:
\begin{align}
    \label{eqn:dec_layers}
    & \tilde{A}_l^D = \overline{\text{attn}}(A_{l-1}^D, A_{l-1}^D, A_{l-1}^D) \\
    & A_l^D = FF\Big(\text{attn}(\tilde{A}_{l}^D, A_{L_E}^E, A_{L_E}^E)\Big) \label{eqn:dec_attn}
\end{align}
Where the first layer attends over the initial encoding of the target sequence: $A_0^D=T$.

\begin{figure}[t!]
    \centering
    \includegraphics[width=0.25\textwidth]{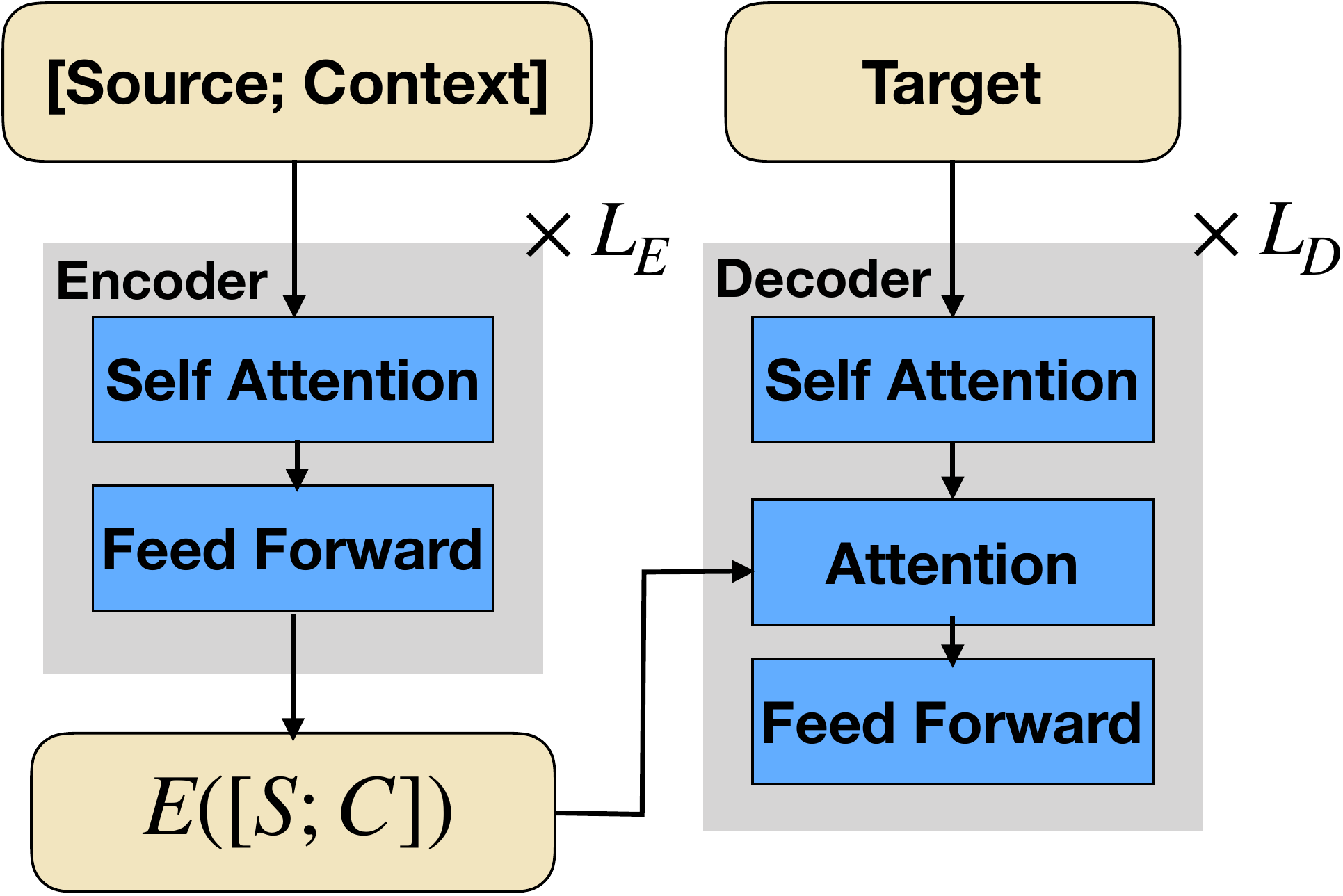}
    \caption{Basic components of a seq2seq Transformer.}
    \label{fig:sequential}
\end{figure}

\paragraph{Adding Context} As mentioned above, the standard Transformer architecture is designed for a single input and output. In order to apply this setting to context-aware tasks, one common approach consists in simply concatenating the query and context with a special separator token between them, and feeding this extended source sequence to the model, as in \cite{wikisum,eli5}. This method is illustrated in Figure~\ref{fig:sequential} and referred to in the rest of this paper as the {\bf{sequential}} approach.

\begin{figure*}
    \begin{subfigure}{0.31\textwidth}
    \includegraphics[width=\textwidth]{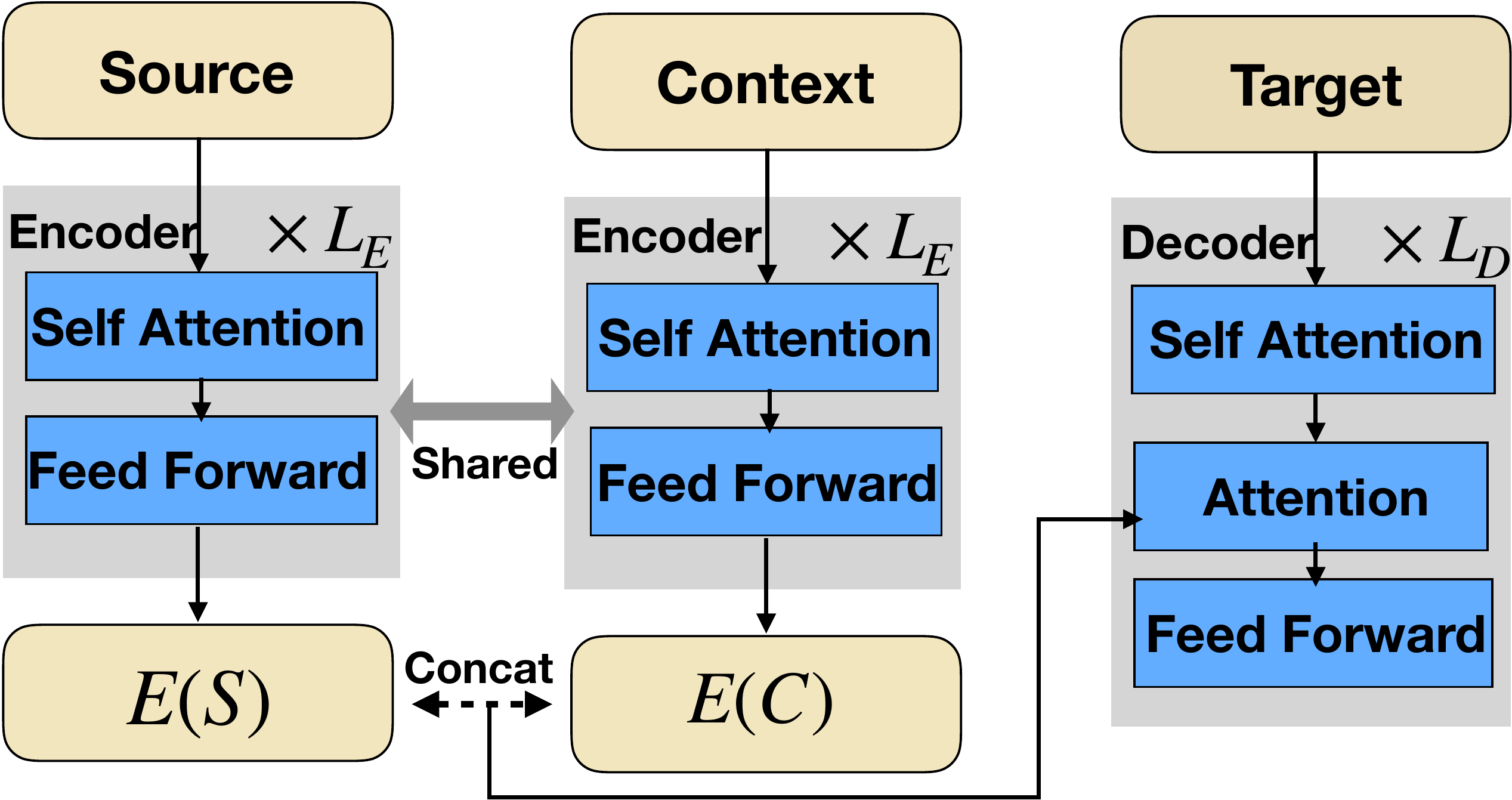}
    \caption{}
    \end{subfigure} \quad
    \begin{subfigure}{0.31\textwidth}
    \includegraphics[width=\textwidth]{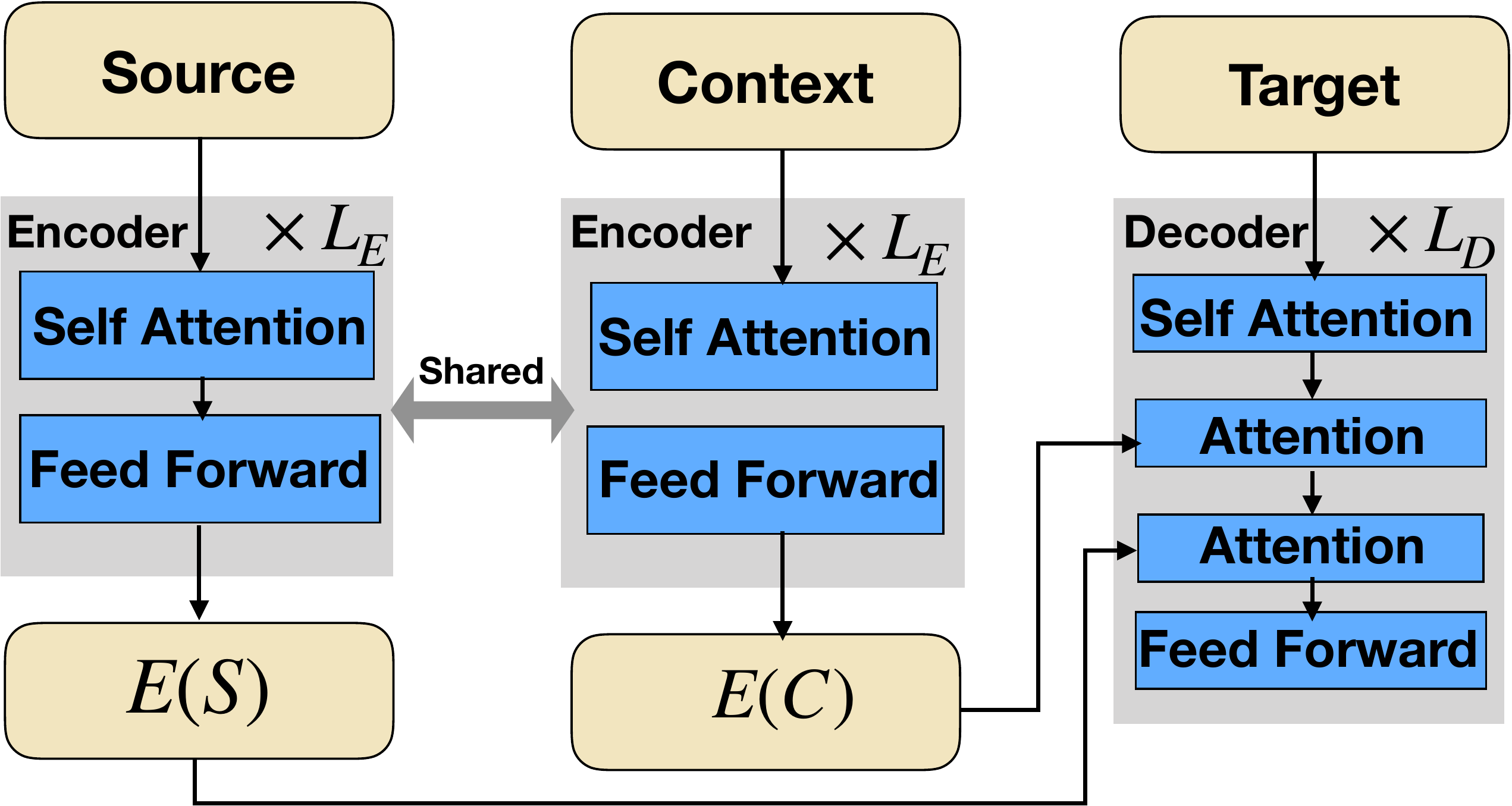}
    \caption{}
    \end{subfigure} \quad
    \begin{subfigure}{0.31\textwidth}
    \includegraphics[width=\textwidth]{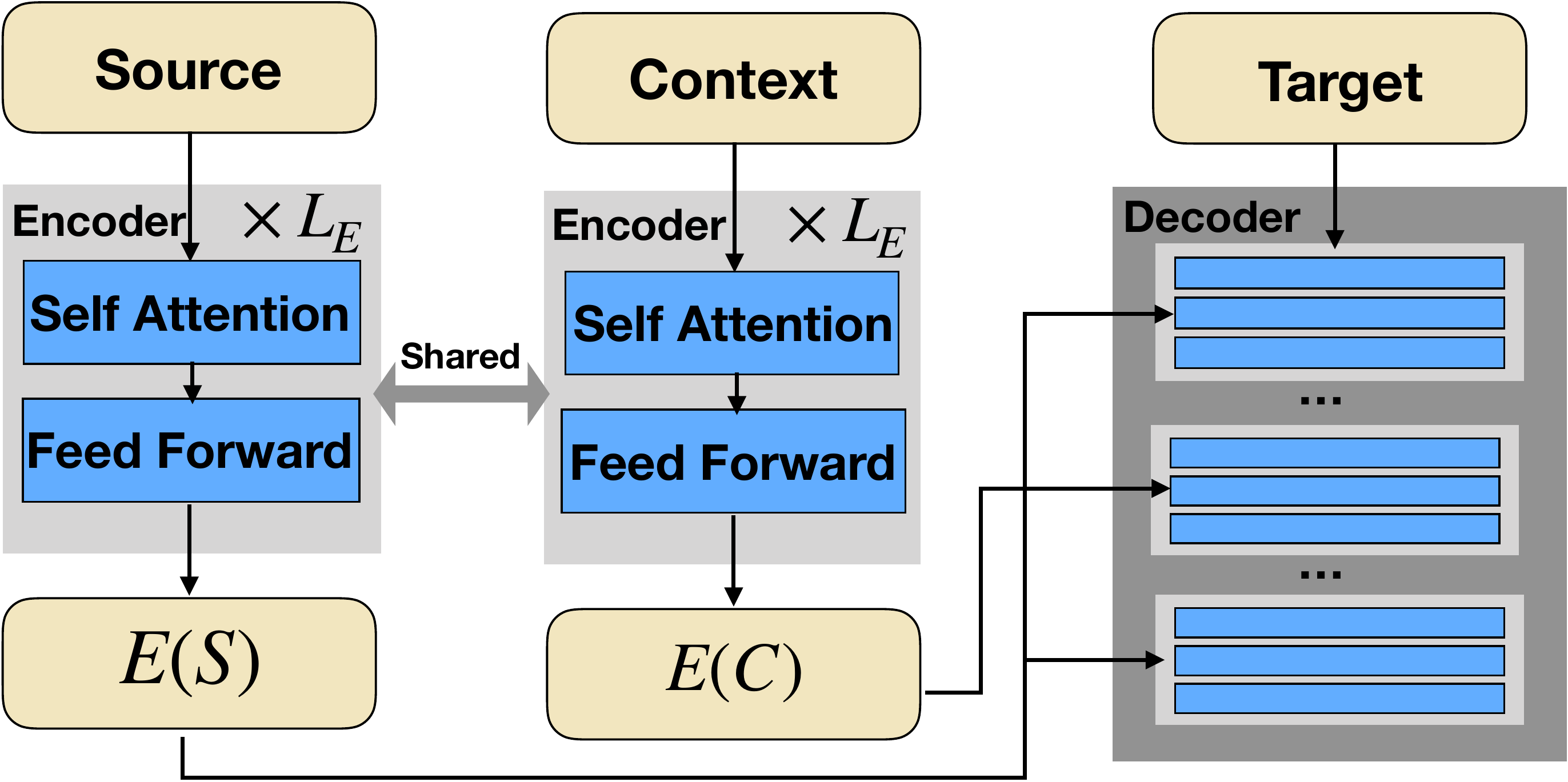}
    \caption{}
    \end{subfigure}
    \caption{Our proposed framework. We only draw one encoder layer for simplicity. For the decoder, we draw one representative layer for (a) and (b) and three for (c). (a) The {\bf concatenate} model. (b) The {\bf alternate} model. (c) The {\bf interleave} model.}
    \label{fig:model}
\end{figure*}

\subsection{Context-Aware Encoding Strategies}

While the {\bf {sequential}} encoding outlined above has led to promising results on some tasks, \cite{wikisum} show that its performance starts decreasing when the input is longer than a few hundred tokens.
One particular limitation is that the decoder treats the source and context in a roughly similar fashion even though they have different roles.
Next, we propose several alternative architectures which separate the encoding of the source and context and combine them in various ways for decoding. The different models are illustrated in Figure~\ref{fig:model}.

\paragraph{Combining Source and Context} In all of the proposed architectures, we apply the encoder separately to $S$ and $C$ and leave the job of combining their information to the decoder. Note that while the source and context have different roles, they are both natural language sequences, so that we choose to use the same encoder for both. We denote as $E(S)$ and $E(C)$ the outputs of the encoder applied to the source and context respectively. We propose three approaches to combining these encodings in the decoder: {\bf concatenate}, {\bf alternate}, and {\bf interleave}.

The most straight-forward way to combine these encodings is the {\bf concatenate} approach: the decoder attention is applied to the concatenation of the source and context encoder, denoted as $[E(S);E(C)]$. Equation~\ref{eqn:dec_attn} then becomes:
\begin{align}
    A_l^D = FF\Big(\text{attn}(\tilde{A}_{l}^D, [E(S);E(C)], [E(S);E(C)])\Big)
\end{align}
Note that while this approach looks similar to the {\bf {sequential}} approach, the {\bf concatenate} removes the cross-attention between the source and context, and the information in both sequences can only be combined by the decoder. Additionally, we can now apply different constraints to the source and context self-attentions as discussed later in this Section.

The {\bf alternate} approach is inspired by the document-level NMT decoder of \cite{doc_nmt}, which adds an extra attention module in each decoder layer to attend over the context $C$ before attending over the context $S$. This extends Equations~\ref{eqn:dec_layers}-\ref{eqn:dec_attn} so that $\forall l \in \{1 \ldots L_D\}$:
\begin{align}
    & \tilde{A}_l^D = \overline{\text{attn}}(A_{l-1}^D, A_{l-1}^D, A_{l-1}^D) \\
    & \hat{A}_l^D = \text{attn}(\tilde{A}_{l}^D, E(C), E(C)) \\
    & A_l^D = FF\Big(\text{attn}(\hat{A}_{l}^D, E(S), E(S))\Big)
\end{align}

Finally, we propose a model to {\bf interleave} the source attention and context attention. In this setting, some decoder layers only attend to the source $S$ and others only attend to the context $C$ after the self-attention module. Let $\mathcal{L}^S$ be the set of layers which attend over the source and $\mathcal{L}^C$ the ones which attend over the context, Equation~\ref{eqn:dec_attn} becomes:
\begin{equation}
   A_l^D = \begin{cases}
    FF\big(\text{attn}(\tilde{A}_{l}^D, E(S), E(S))\big) & \text{if}\ l \in \mathcal{L}^S \\
    FF\big(\text{attn}(\tilde{A}_{l}^D, E(C), E(C))\big) & \text{if}\ l \in \mathcal{L}^C
   \end{cases}
\end{equation}

Now that we have presented different ways of combining the source and context, let us consider how else we can take advantage of the separate encoding.

\paragraph{Focused Context Attention} Although the $S$ and $C$ representations are computed with the same set of encoder parameters, we apply two modifications to the encoder self-attention when processing $C$, to encourage better utilization of the often long and noisy context.
First, we help the model ignore some of the noise in the context by adding a {\bf temperature} term to the regular self-attention function, so that Equation~\ref{eqn:attn_score} becomes:
\begin{align}
        \alpha' = \text{softmax} (\tau \cdot (Q P_Q) (K P_K)^T ) \label{eqn:attn_window}
\end{align}
where $\tau$ is the temperature hyper-parameter that controls the peakiness of the self-attention distribution for the context $C$. Intuitively, a bigger $\tau$ would skew the distribution towards words with higher scores, thus making the model focus on more important information when encoding the relatively long context $C$.

Secondly, the context $C$ is often much longer than $S$, which might make it difficult to learn a shared set of encoder parameters. Inspired by the local attention in \cite{dot_prod_attention}, we add a {\bf localized window} for the self-attention scores in $C$ to remedy this: 
\begin{align}
       \alpha''_{ij} = \alpha'_{ij} \cdot \text{exp}(-\frac{(i-j)^2}{\sigma^2}) \cdot c_j  \label{eqn:attn_temp}
\end{align}
where $\sigma$ is a soft attention window size centered around the current word $c_i$. Thus, the effective size of the encoder self-attention window is about the same for $S$ and $C$.

\subsection{Data Augmentation}
Finally, since our model encodes $S$ and $C$ separately, we can easily perform data augmentation by randomly removing or swapping any of the $S$, $C$ and $T$ sequences. Specifically, we propose to randomly change the decoding task for some of the examples. With probability $p_{S\text{-}T}$, we randomly remove $C$ from the input so that the model learns to generate $T$ using $S$ only, which we believe will help the model produce outputs which are relevant to the source. Then, with probability $p_{S\text{-}C}$, we ask the model to predict the context $C$ conditioned on $S$. This has the double advantage of letting the encoder learn which parts of a query are related to the support documents and acting as a form of language model pre-training for the decoder, since the context is typically longer than the standard output sequence.
\section{\label{sec:exp}Experiments}

We validate our approach on three different tasks which involve making use of a context document: long form question answering (LFQA), knowledge-grounded dialogue, and document-level machine translation. Table~\ref{tab:data_statistics} presents basic statistics for all three datasets. We first describe the three task settings along with relevant baselines, then compare these to the results of our approach, and finally present additional analysis to investigate how each of our proposed method's components contributes to its performance gains.

\subsection{Datasets and Baselines}
 \begin{table}[]
     \centering
     \begin{tabular}{l|cccc}
     \toprule
    Dataset & Size  & $|S|$ &  $|C|$ & $|T|$ \\
     \midrule
    ELI5    & 272K  & 42    & 858   & 131 \\
    Wizards & 74K   & 28    & 81    & 24 \\
    Doc-MT  & 220K  & 24    & 80    & 23 \\
    \bottomrule
     \end{tabular}
     \caption{Dataset statistics for long form QA (ELI5), knowledge-grounded dialogue (Wizards of Wikipedia), and document-level MT.}
     \label{tab:data_statistics}
 \end{table}

\paragraph{ELI5: Long Form Question Answering}
We first apply our approach to the recently published ELI5 dataset~\cite{eli5} for LFQA. The dataset consists of 272,000 complex questions and answer pairs, along with supporting documents created by gathering and concatenating passages from CommonCrawl web pages which are relevant to the question. The questions are typically elaborate and require paragraph-length answers (42 and 131 tokens long on average respectively), with context documents averaging 858 words. We follow~\cite{eli5} in reporting \textsc{rouge-1}, \textsc{rouge-2}, and \textsc{rouge-l}\footnote{We use the open-source \textsc{rouge} implementation of \url{https://github.com/google-research/google-research} as it is closer to the original version of~\cite{rouge}, hence the difference in reported numbers for the baseline. Our re-implementation does replicate the results of \cite{eli5} within .01 using their evaluation script.} between the model generation and gold answer, as well as gold answer perplexity (PPL).

\begin{table}[!t]
    \centering
    \begin{tabular}{l|lll|l}
    \toprule
    Model & R-1 & R-2 & R-L & PPL \\
    \midrule
    Sequential & 22.14 & 4.24 & 13.83 & 50.23 \\
    Multitask & 22.48 & 4.63 & 14.35 & 31.66  \\
    Multi + Sequential & 23.02 & \textbf{4.80} & 14.52 & \textbf{30.63} \\
    \midrule
    Interleave & 23.13 & 4.62 & 14.41 & 36.10  \\ 
    Concatenate & 22.86 & 4.54 & 14.33 & 36.37 \\ 
    Alternate  & 22.79 & 4.54 & 14.24 & 36.95 \\ 
    \midrule
    Multi + Interleave  & \textbf{23.32} & \textbf{4.79} & \textbf{14.63} & 30.80  \\
    \bottomrule
    \end{tabular}
    \caption{Results on the ELI5 dataset.}
    \label{tab:eli5_results}
%   MultiT-3 & 22.60 & 4.59 & 14.32 & 35.35  \\
\end{table}

We report results for the {\bf sequential} baseline and the author's \textbf{multitask} system. The latter is trained on 10 different tasks, including predicting using all combinations of $S$, $C$, and $T$ as input and output, as well as a BERT-like~\cite{bert} masked word prediction task. This leads to a significant improvement in both \textsc{rouge} and perplexity. Finally, since the model is only tested on the setting where the model predicts the target $T$ given the source $S$ and context $C$, we fine-tune a \textbf{multitask}-trained model using both the standard {\bf sequential} approach and our {\bf interleave} setting. Results for these systems, as well as for all our proposed encoding strategies, are presented in Table~\ref{tab:eli5_results}.

\paragraph{Wizards of Wikipedia: Knowledge Grounded Dialogue} Next, we test our framework on the Wizards of Wikipedia dataset~\cite{wizard}. We consider the gold-knowledge setting, where the system is provided both with a dialogue history and with a sentence or short passage from Wikipedia, and required to provide a next dialogue utterance using the information contained in the passage. Given the relative sizes of both inputs (see Table~\ref{tab:data_statistics}), we choose to use the Wikipedia fact as the source $S$, and the longer dialogue history as context $C$. We re-implemented the authors' End2End system, which is most similar to our {\bf concatenate} setting without the focused context attention or data augmentation, and report the unigram F1 score between the model generation and the gold response\footnote{We use the F1 calculation script from Parlai~(\url{https://parl.ai/}).} and the perplexity in Table~\ref{tab:wizard_results}.

% We compare with multiple methods: 1) Regular: again this is a regular seq2seq transformer that uses the concatenation of the knowledge $S$ and the dialogue history $C$ as the input, and generates a response $T$; 2) End2End: this is the best performing method in \cite{wizard}. It essentially follows the encoder and decoder architecture from the concatenate setting in our framework, but without the data dropout and special attention strategies. The model uses a single encoder which separately encodes the knowledge and dialogue history; the two encodings are concatenated before feeding into the standard transformer decoder. 
\paragraph{Document Level Machine Translation} Finally we test our method on a document-level MT task. We use the IWSLT Fr-En dataset, which contains 220K parallel sentences from 1,824 TED talk documents~\cite{iwst}. Following the setting of \cite{doc_nmt}, we translate each source sentence $S$ in the French document into the corresponding sentence $T$ in the English document, using the previous 2 French sentences as the document context $C$. We use the IWSLT 2010 development and test sets, and report the BLEU score to measure the quality of the translation model. \cite{doc_nmt} propose a Transformer architecture which first encodes the context, uses the context representation to encode the source, then uses both of the context and source encoding when decoding by alternating attention modules similarly to our {\bf alternate} architecture. We re-implement their model and refer to it as {\bf doc-NMT}, and present its BLEU numbers as well as the {\bf sequential} baseline (with and without document context) and our proposed architectures in Table~\ref{tab:nmt_results}.

% We implement and compare with several different methods as competitive baselines: 1) Regular (w/o ctx): this is simply the standard transformer model using only sentence level data. We compare with this method because prior works in document-level NMT found that adding context $C$ to an NMT model can sometimes underperform sentence-level models because of overfitting; 2) Regular (w ctx):  we use a regular seq2seq transformer model, which takes the concatenation of source $S$ and context sentences $C$ as a single text input, and generate the translation $T$; 3) doc-NMT: the document-level NMT model proposed by . It encodes the context with additional parameters, and integrates the context encoding into both the encoder and the decoder using extra multi-head attention layers. 

\paragraph{Additional Implementation Details}  We implement all our models and baselines on top of the Fairseq toolkit~\cite{fairseq} using Byte Pair Encoding (BPE)~\cite{bpe}. All models have 6 encoding and 6 decoding layers; the {\bf interleave} architecture attends to the source at layers 1, 2, 5, and 6, and attends to the context at layers 3 and 4. For ELI5, we mostly follow the hyper-parameters in \cite{eli5} and use their BPE vocabulary. For decoding, we did not a set minimum generation length as in \cite{eli5}, but use a length penalty of 1.5, as we found that this decoding strategy results in fewer repetitions in generation. For document-level MT, we follow the approach of \cite{doc_nmt,hier_doc_nmt,tu_cache} in first training the model to convergence on sentence pairs only, then fine-tuning using the context.

\begin{table}[!t]
    \centering
    \begin{tabular}{l|ll}
    \toprule
    Model & F1 & PPL \\
    \midrule
    Sequential & 34.61 & 22.89 \\
    End2End~\cite{wizard} & 35.50 & 23.10 \\
    End2End~(Our implementation) & 35.31 & 21.53 \\
    \midrule
    Interleave & \textbf{35.69} & 19.73 \\
    Concatenate & 35.60 & 19.63 \\ 
    Alternate  & 35.62 & \textbf{19.62} \\ 
    \bottomrule
    \end{tabular}
    \caption{Results on the Wizard of Wikipedia dataset.}
    \label{tab:wizard_results}
    \vspace{.1in}
% \end{table}
% \begin{table}[]
%    \centering
    \begin{tabular}{l|l}
    \toprule
    Model & BLEU \\
    \midrule
    Sequential~(w/o ctx)~\cite{doc_nmt} & 35.17 \\
    doc-NMT~\cite{doc_nmt} & 36.04 \\
    \midrule 
    Sequential~(w/o ctx) & 36.47 \\
    Sequential~(w ctx) & 36.85 \\
    doc-NMT~(our implementation) & 36.61 \\
    \midrule
    Interleave  & $\textbf{37.30}$ \\
    Concatenate & 37.08 \\ 
    Alternate   & 36.81 \\ 
    \bottomrule
    \end{tabular}
    \caption{BLEU scores for the document-level MT task.
    }
    \label{tab:nmt_results}
\end{table}

\subsection{Summary of Results}

Comparing the results from Tables~\ref{tab:eli5_results}, \ref{tab:wizard_results}, and \ref{tab:nmt_results}, a few trends emerge. First, all of our proposed approaches always out-perform the standard {\bf sequential} seq2seq training as well as the {\bf end2end} and {\bf doc-NMT} models on the Wizards of Wikipedia and document-level MT respectively. These results validate the importance of the data augmentation method, as well as the separate source and context encoding with focused context attention. Between the three proposed architectures, {\bf interleave} consistently outperforms {\bf concatenate} and {\bf alternate}, with a slightly higher perplexity on Wizards of Wikipedia. This is especially notable as it is the architecture with the least number of operations, since each layer only attends to one of the sources.

On the ELI5 dataset, {\bf interleave} still lags a little behind {\bf multitask} in terms of perplexity but achieves better \textsc{rouge}, which, according to \cite{eli5}, correlates better with human judgements. Note also that while our data augmentation method randomly samples a different input and output setting in up to 50\% of cases, the {\bf multitask} setting has to go through 10 different versions of each example, leading to much slower convergence (24 hours for {\bf interleave}, 5 days for {\bf multitask}). Further, since the {\bf interleave} architecture uses the same set of parameters as the standard encoder-decoder transformer, we can easily fine-tune a trained multitask model using our framework. Table~\ref{tab:eli5_results} shows that this approach significantly improves over the regular {\bf multitask} (as well as over fine-tuning with the {\bf sequential} setting), indicating that our approach and the multitask training of \cite{eli5} are complementary.

The difference between the three architectures is least pronounced on the Wizard of Wikipedia dataset. One significant difference between this and the other tasks is that whereas the source has a larger role in guiding the answer in question answering or sentence-level translation with document-level context, the chat history context may be more relevant than the supporting fact in knowledge-grounded dialogue. Finally, we note that in the document-level MT task, the {\bf sequential} setting is actually the strongest baseline, performing better than {\bf doc-NMT}. Still, our {\bf interleave} model retains an advantage over that approach as shown in Table~\ref{tab:nmt_results}.

\begin{table}[!t]
    \centering
    \begin{tabular}{l|ccc|c}
    \toprule
    Model           & R-1 & R-2 & R-L & PPL \\
    \midrule
    Interleave      & 23.13 & 4.62 & 14.41 & 36.10  \\ 
    \midrule
    -Attn Window    & 22.68 & 4.55 & 14.24 & 36.19 \\ 
    -Attn Temperature     & 23.00 & 4.62 & 14.32 & 37.06 \\ 
    -Data Augmentation      & 22.40 & 4.30 & 13.91 & 48.61 \\ 
    \midrule
    Sequential      & 22.14 & 4.24 & 13.83 & 50.23 \\
    \bottomrule
    \end{tabular}
    \caption{Ablation results for the ELI5 task.}
    \label{tab:eli5_ablate}
    \vspace{.1in}
    \centering
    \setlength\tabcolsep{3pt}
    \begin{tabular}{l|cc}
    \toprule
    Model & F1 & PPL \\
    \midrule
    %Concat. & 35.60 & 19.63 \\ 
    Interleave & 35.69 & 19.73 \\ 
    \midrule
    %-Attn Win.    & 35.34 & 19.71 \\
    %-Attn Tem.     & 35.32 & 19.76 \\
    %-Data Aug.      & 35.50 & 21.44 \\
    -Attn Win.    & 35.45 & 19.74 \\
    -Attn Tem.     & 35.56 & 19.71 \\
    -Data Aug.      & 35.17 & 21.37 \\
    \midrule
    Sequential      & 34.61 & 22.89 \\
    \bottomrule
    \end{tabular}
    \qquad \quad
    \begin{tabular}{l|c}
    \toprule
    Model & BLEU \\
    \midrule
    Interleave  & 37.30 \\
    \midrule
    -Attn Win.    & 37.21 \\
    -Attn Tem.     & 37.16 \\
    -Data Aug.      & 37.01 \\
    \midrule
    Sequential      & 36.85 \\
    \bottomrule
    \end{tabular}
     \caption{Ablation results for the  Wizard of Wikipedia {\bf (left)} and document-level NMT {\bf (right)} tasks.}
     \label{tab:wizard_nmt_ablate}
\end{table}

\subsection{Further Analysis}

We have shown that our proposed approach leads to consistent improvement over strong baselines in a variety of context-aware tasks. In the rest of this Section, we run further experiments to better understand where the performance gains come from.

\paragraph{Ablation Studies}
We start with ablation experiments to determine the relative importance of the soft attention window (as described in Equation~\ref{eqn:attn_window}), attention temperature (Equation~\ref{eqn:attn_temp}), and data augmentation. We remove one component at a time from the training process for the {\bf interleave} architecture for all three tasks, and provide results in Tables~\ref{tab:eli5_ablate} and \ref{tab:wizard_nmt_ablate}. Overall, we can see that each of the components contributes to the model's success, with the data augmentation having the most impact, especially on ELI5. The attention window also seems more important on Long Form QA, which we attribute to the fact that its contexts are an order of magnitude larger than in the document-level MT and knowledge-grounded dialogue tasks.

\begin{table}[]
    \centering
    \begin{tabular}{lll}
    \toprule
     Model & bw PPL & $U_{\text{ctx}}$\\
    \midrule
    Sequential & 317.70 & 31.11 \\
    Multitask & 255.43 & 32.54 \\
    \midrule
    Interleave & \textbf{253.24} & \textbf{34.03} \\
    Concatenate & 299.64 & 33.37 \\
    Alternate & 298.26 & 32.58 \\
    \bottomrule
    \end{tabular}
    \caption{Backward PPL of the question given the generation.}
    \label{tab:bw_ppl}
\end{table}

\begin{table}[]
    \centering
    \begin{tabular}{l|lll}
    \toprule
       Method  & $S_{\text{attn-}C}$ & $C_{\text{attn-}S}$ & $\text{Win}_{\text{attn}}$ \\
    \midrule
       Sequential  & 0.736 & 0.292 & 0.170 \\
       Multitask  &  0.480 & 0.310 & 0.274 \\
    \bottomrule
    \end{tabular}
    \caption{Statistics of the encoder attention of the baseline Transformer models for ELI5.}
    \label{tab:attn_analysis}
\end{table}

\begin{table*}[]
    \centering
    \resizebox{!}{0.15\textwidth}{
    \begin{tabular}{l|p{5cm}|p{5cm}|p{5cm}|p{5cm}}
    \toprule
    Task & Source & Context & Baseline & Interleave \\
    \midrule
    Dialogue & sight words, often also called high frequency sight words, are commonly used words that...so that they can automatically recognize these words in print without having to use any strategies to decode. & reading\ldots my mom did the same. i started reading sight words when i was about 3.5 years old. \textcolor{red}{what is your favorite genre and/or book?} & \textcolor{red}{that's a good idea.} i love reading because it is often called high frequency words, which is commonly used for young children that are encouraged to memorize as a whole sight. & \textcolor{red}{i'm not sure}, but i do know that reading is often called high frequency words, which is commonly used words that young children are encouraged to memorize as a whole sight. \\
    \midrule
   MT & \textcolor{red}{l\`a}, ils comprenaient l'importance du prototypage. & nous avons refait l'exercice avec les m\^emes \'etudiants. que pensez-vous qu'il est \textcolor{red}{alors} arriv\'e ? & \textcolor{red}{there} they understood the importance of prototyping. & \textcolor{red}{now}, they understood the importance of prototyping. \\
    \midrule
   QA & \textcolor{red}{Why} do some people prefer sleeping on their sides while others prefer sleeping on their stomach or back?  &  People who have obstructed breathing problems will find themselves sleeping in different positions than others\ldots But some sleep positions can be better for you than others, depending on what kind of ailment you may have\ldots & Some people prefer sleeping on their sides, while others prefer to sleep on their back. & Some people prefer to sleep on their sides \textcolor{red}{because} it's easier to fall asleep on their back than on their side. Some people are more comfortable with their sides than others. \\
    \bottomrule
    \end{tabular}}
    \caption{Example generations of our model and the best baseline for each task.}
    \label{tab:example}
\end{table*}

\paragraph{Effect on Source and Context Utilization} Next, we evaluate how much our approach affects the model's utilization of the source $S$ and context $C$. We propose two metrics to measure these effects, focusing on the long form QA task for this part of the analysis. 

First, we report the \textbf{backward perplexity}~(bw PPL) of the source question $S$ given the generated answer $\hat{T}$. To measure the backward perplexity, we first train a seq2seq Transformer model on the training data to predict an example's source (question) given the gold answer. We then generate answers on the full test set for each of the considered models, and report the perplexity of the questions conditioned on these generations under the trained model. Intuitively, the more relevant to a question a generated answer is, the lower the perplexity should be.

Next, we define the \textbf{context use percentage}~($U_{\text{ctx}}$) as a measure of how well the model utilizes the relevant information from the context. We run a part-of-speech tagger on the context, extract the set of nouns from the context which are present in the gold answer, and compute the percentage of these nouns that are present in the generated output: 
\begin{align}
    U_{\text{ctx}} = \frac{ |N(\text{ctx}) \cap N(\text{gold}) \cap N(\text{generate})| }{ | N(\text{ctx}) \cap N(\text{gold})| } \times 100
\end{align}
Essentially, the context percentage measures how much of the ``useful'' information present in the context is utilized by the model to generate the target. Higher $U_{\text{ctx}}$ indicates that the model is better at identifying an using the relevant information in $C$.

The backward perplexity and context usage percentage of the baselines as well as the {\bf interleave}, {\bf concatenate}, and {\bf alternate} methods for the ELI5 dataset are given in Table \ref{tab:bw_ppl}. First, we note that the {\bf multitask} model performs better than {\bf sequential} on both metrics, which is coherent with the findings from \cite{eli5}'s human evaluation of the model outputs, and validates the use of the proposed metrics. While all of our models perform better than the {\bf sequential} baseline in terms of backward perplexity and than both baselines on context use percentage, {\bf  interleave} is ahead of all of the other settings on both fronts, confirming the \textsc{rouge} numbers of Table~\ref{tab:eli5_results}. Additionally, the difference in $U_{\text{ctx}}$ appears to be the more significant of the two, indicating that our method especially improves over {\bf multitask} in terms of making good use of the context.

\paragraph{Effect of Encoder Attention Constraints} We introduced two constraints to the self-attention of our model encoders. First, while the encoder in the {\bf sequential} setting represents the source and context as a single sequence and can thus have cross-attention between $S$ and $C$, all of our architectures encode both separately, preventing this from happening. Secondly, we introduced a soft {\bf localized window} in the context encoder, which means that each token encoding depends on fewer neighbors than in the {\bf sequential} setting.

We investigate how these restrictions affect our models by analysing the encoder attention patterns for the {\bf sequential} and {\bf multitask} baselines for long form QA in Table~\ref{tab:attn_analysis}. Specifically, we measure how much of the attention weights used to compute the source token representations is spent on context tokens ($S_{\text{attn-}C}$), and how much the encoding of context tokens attends over the source tokens ($C_{\text{attn-}S}$), as well as how much of the self-attention falls within a localized window of 40 tokens ($\text{Win}_{\text{attn}}$).

At first glance, these constraints appear to significantly limit the model's representation power, especially when looking at the {\bf sequential} setting where 74\% of the source token encoding attention is spread over context tokens and only 17\% of the attention in general falls within a 40-tokens local window. However, we note that {\bf multitask} training, which performs better on all measures, significantly reduces the source-to-context cross-attention ($S_{\text{attn-}C}$ dropping from 74\% to 48\%) and has a more strongly localized attention ($\text{Win}_{\text{attn}}$ rises from 17\% to 27\%). This indicates that the proposed constraints act as a positive inductive bias rather than as a limitation of the model's capacity.

\paragraph{Qualitative Analysis} Finally, we provide some representative example outputs for our model and the best baseline for each of the tasks. The source, context, and both generations are shown in Table \ref{tab:example}. While both generations for the Wizards of Wikipedia example overly rely on the source, our model does provide an answer which is more fluent given the context (answering ``what is your favorite genre'' with ``'I don't know'' rather than ``that's a good idea''). In the document-level MT example, the system needs to translate the word ``l\`a'', which can mean either ``now'' or ``there''. Our system identifies the word ``alors'' (``then'') in the context as a cue for this translation, and correctly chooses the temporal over the geographical meaning. Lastly, in the ELI5 example, our model makes better use of the question word ``why'' and provides an actual explanation instead of simply rephrasing the question as an assertion.

\section{\label{sec:conclusion}Conclusion}

In this work, we have examined the general problem of training models for seq2seq tasks that conditioned on both a source sequence and a context document. We first identified some challenges specific to this family of problems, then proposed a new model architecture which separates source and context encoding and interleaves source and context attention when decoding, as well as a data augmentation strategy to address these challenges. We have shown that our approach yields consistent overall improvements over strong baselines on three different tasks, and conducted an extensive investigation of the mechanisms underlying these improvements, confirming that the proposed architecture and training procedure lead the models to make more efficient use of both the source and context sequences.

\newpage

\bibliography{iclr2020_conference}
\bibliographystyle{aaai}

\newpage

\appendix

\section{Implementation Details}
Our code is implemented on top of the Fairseq toolkit. For our method, we search over the attention temperature of $\tau = \{2, 4, 32\}$ and window size of $\sigma = \{40, 80, 100\}$. For data augmentation, we use $\{p_{S\text{-}T} = 0.3, p_{S\text{-}C}=0.2\}$ or $\{p_{S\text{-}T} = 0.2, p_{S\text{-}C}=0.1\}$.

\subsection{Long-form QA}
Here are the detailed hyperparameters for our method:
\begin{itemize}
    \item We follow the hyperparameter settings for the transformer model in \cite{eli5}.  The transformer model has 6 layers, 16 attention heads. The word embedding is set to 1024, and the feed-forward layer has the dimension of 4096. 
    \item We use the Adam optimizer with learning rate of $10^{-4}$. The warmup steps is set to 4K, and the initial warmup learning rate is $10^{-7}$.
    \item We use Byte Pair Encoding~(BPE) to process the inputs and outputs, using the BPE codes provided by \cite{eli5}. The ovearall vocabulary size is about 53K. 
    \item We share all the embeddings in the transformer model.
    \item For decoding, we use a beam size of 5, and length penalty of 1.5. We set the maximum length to be 500. We also make sure that the model does not have repeated 3-gram.
    \item We use $\tau=32$, $\sigma=40$, $\{p_{S\text{-}T} = 0.3, p_{S\text{-}C}=0.2\}$ as hyperparameters.
\end{itemize}

\subsection{Knowledge-grounded Dialogue}
\begin{itemize}
    \item We mostly follow the hyperparameters of the transformer model for \cite{wizard} in Parlai~(\url{parl.ai}). We use a transformer model of 5 layers and 2 attention heads. The embedding size is set to 256, and the dimension of the feed-forward layer is set to 512.
    \item We use the Adam optimizer with learning rate of $5 \times 10^{-4}$. The warmup steps is set to 5K, and the initial warmup learning rate is $10^{-7}$.
    \item We use Byte Pair Encoding~(BPE) to process the inputs and outputs, using the BPE codes provided by \cite{wizard}. We also lowercase all the data, following the settings in \cite{wizard}. The final vocab size is about 26K. We find that using the same vocabulary is important to replicate the PPL metric.
    \item We share all the embeddings in the transformer model.
    \item We use $\tau=4$, $\sigma=80$, $\{p_{S\text{-}T} = 0.2, p_{S\text{-}C}=0.1\}$ as hyperparameters.
\end{itemize}

\subsection{Document-level NMT}
\begin{itemize}
    \item We mostly follow the hyperparameters of the transformer model in \cite{doc_nmt}. The transformer model has 6 layers and 8 attention heads. We use an embedding size of 512, and the feed-forward layer has size of 1024.
    \item We use the Adam optimizer with learning rate of $10^{-4}$. The warmup steps is set to 4K, and the initial warmup learning rate is $10^{-7}$.
    \item We use BPE to process the input and output. We directly use the processed data from \cite{doc_nmt}.
    \item We share all the embeddings in the transformer model.
    \item We use $\tau=2$, $\sigma=100$, $\{p_{S\text{-}T} = 0.3, p_{S\text{-}C}=0.2\}$ as hyperparameters.
\end{itemize}

\section{Backward PPL model}
We train a backward model to analyze how well the generation is conditioned on the source $S$. Here we provide the details of the backward model we used.

We use a vanilla transformer model that takes the target $T$ as input, and the source $S$ as output. The hyperparameters of the backward transformer model are the same with the forward model described in the previous section.

\section{Effect of Data Augmentation Components}

\begin{table}[]
 \centering
\begin{tabular}{lll}
 \toprule
  Model & bw PPL & $U_{\text{ctx}}$\\
 \midrule
 Interleave & 253.24 & 34.03 \\
 \midrule
 no $S$-$T$ & 482.25 & 35.56 \\
 no $S$-$C$ & 453.78 & 31.59 \\
 \bottomrule
 \end{tabular}
 \caption{Backward PPL on Source}
 \label{tab:data_effect}
\end{table}

In this Section we analyze the different effects of the two types of data augmentations at targeting the challenges facing the query-context conditioned generation. We remove either of the $S$-$T$ or the $S$-$C$ data augmentations, and check how that affects the model's ability at conditioning on the query and utilizing the document. We check the quantitative measures of bw PPL and $U_{ctx}$ on the ELI5 QA task after removing either of the data augmentations. The results are listed in Table \ref{tab:data_effect}. Removing the $S$-$T$ data augmentation hurts the bw PPL the most, which indicates that sampling $S$-$T$ data helps the model to generate targets that better adress the query $S$. On the other hand, removing the $S$-$C$ data augmentation significantly reduces the context utilization $U_{\text{ctx}}$, which indicates that sampling $S$-$C$ is critical to utilizing the context effectively.

\end{document}